\documentclass{article}
\usepackage{spconf,amsmath,graphicx,hyperref}
\usepackage{newfloat}
\usepackage{listings}
\usepackage{url}            
\usepackage{booktabs}       
\usepackage{amsfonts}       
\usepackage{nicefrac}       
\usepackage{microtype}      
\usepackage{xcolor}         
\usepackage[numbers]{natbib}
\usepackage{amsmath}
\usepackage{algorithm}
\usepackage{algorithmic}   
\usepackage{graphicx}       
\usepackage{subcaption}   

\usepackage{booktabs}        
\usepackage{siunitx}         
\usepackage{threeparttable}  
\usepackage{adjustbox}  
\usepackage{multirow}
\usepackage{multicol}

\usepackage[table]{xcolor}
\usepackage{adjustbox}

\usepackage{amssymb}    

\title{Mitigating Hallucinations in Large Language Models Via Decoder Layer Skipping}
\name{Hanze Li, Jinhao You, Yichen Guo, Kai Tang, Shuangyang Xie, Xiande Huang\sthanks{Corresponding author}}
\vspace{-0.3cm}
\address{$^{*}$De Artificial Intelligence Lab
         }
\begin{document}
\ninept
\maketitle

\begin{abstract}
Large Language Models (LLMs) have achieved strong performance across diverse natural language tasks, yet their outputs often suffer from hallucinations—content that is misaligned with factual information. In this work, we conduct a comprehensive layer-wise analysis of the decoding process and reveal that hallucinations tend to originate from deeper decoder layers. To address this issue, we introduce \textbf{DeLask} (\textbf{De}coder \textbf{La}yer \textbf{Sk}ipping), a novel decoding framework that dynamically skips layers prone to producing hallucinations. DeLask leverages the theoretical insight that the forward computation of an $L$-layer Transformer is conditionally equivalent to $L$ steps of gradient descent. We define a \emph{driftance value} by computing the cosine similarity between gradients derived from consecutive decoder steps, identifying problematic layers when the descent direction reverses. Rather than discarding such layers entirely, DeLask partially aggregates their hidden states with preceding layers, thereby preserving consistency while suppressing erroneous signals. Extensive experiments across diverse LLMs and benchmarks demonstrate that DeLask consistently mitigates hallucinations and enhances overall reliability, providing a lightweight and generalizable decoding framework for improving the robustness of large-scale language models.

\end{abstract}


\section{Introduction}
Large Language Models (LLMs) have exhibited remarkable performance in a broad range of textual information processing tasks, showing strong capabilities in reasoning, translation, text summarization, and interactive systems \citep{touvron2023llama2, grattafiori2024llama}. Although recent LLMs have become more capable when incorporating many new functions, they are still struggling with hallucinations, which means that the content generated by LLMs is frequently unaligned with factual information \citep{ji2023survey}. Hallucinations severely limit the utilization of LLMs in high-stakes areas, such as clinical or legal settings \citep{chuang2023dola}, where rigorous trustworthiness and reliabilities of LLMs are required.

\begin{figure*}[htbp]
    \centering
    \begin{subfigure}[b]{0.55\linewidth}
        \centering
        \includegraphics[width=\linewidth]{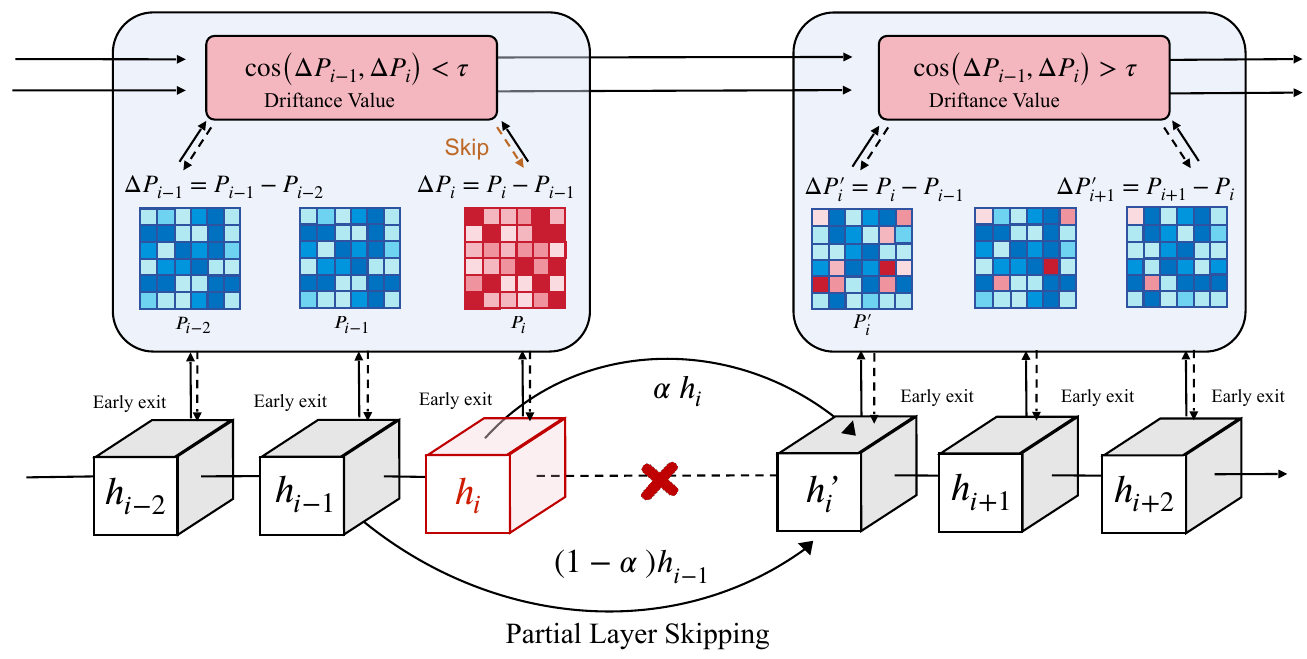}
        \caption{}
        \label{fig:struc}
    \end{subfigure}
    \begin{subfigure}[b]{0.41\linewidth}
        \centering
        \includegraphics[width=\linewidth]{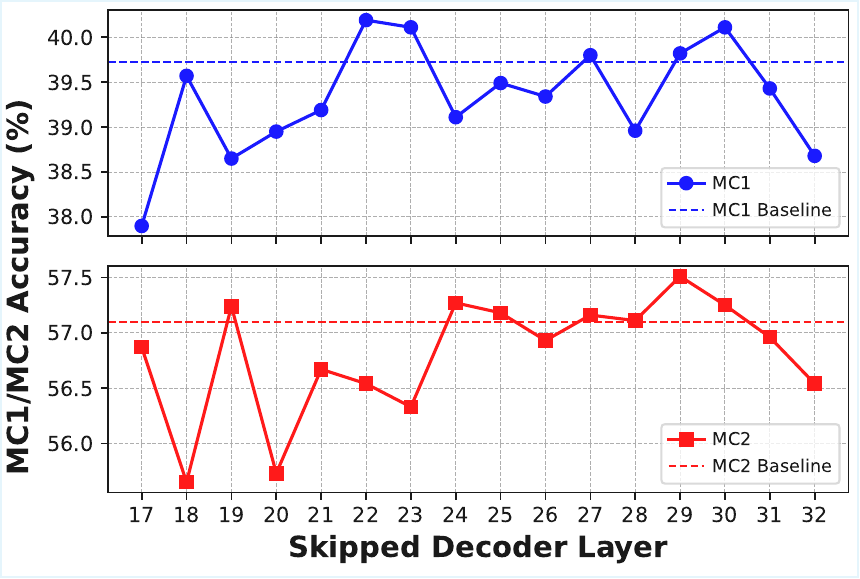}
        \caption{}
        \label{fig:intro_acc}
    \end{subfigure}
    \caption{(a) \textbf{Overall Structure of DeLask}: We first adopt an early-exit strategy that projects the hidden states from each decoder layer onto the final linear layer, producing a probability distribution over the vocabulary for every layer. Next, we compute the difference of the distributions between adjacent layers. From each pairwise difference, we derive a driftance value, defined as the cosine similarity between successive distribution differences. A strongly negative driftance value suggests that the corresponding layer is more susceptible to hallucinations. Once such layers are identified, we partially skip them to mitigate their adverse effect on overall model performance. (b) \textbf{Effects of Partial Layer Skipping on TruthfulQA Dataset}: It demonstrates the influence of layer skipping on model performance, clearly indicating that skipping certain decoder layers enhances the accuracy.}
    \label{fig:cifar_acc}
    
\end{figure*}

Several studies have investigated hallucinations in the decoding process of LLM, identifying suspicious causes of the issue, such as errors in training data \citep{mckenna2023sources} and  
the deviation in model parameter-fitting \cite{gallegos2024bias}. Recent methods to mitigate hallucinations include fine-tuning LLMs using either synthetically generated \cite{song2024rag} or real-world \cite{zimmerman2024two} datasets specifically designed to correct hallucinated outputs, as well as employing Reinforcement Learning from Human Feedback (RLHF) techniques \citep{yu2024rlhf}. However, these hallucination-mitigating approaches require retraining the model, resulting in substantial computational resource consumption. Alternatively, some methods mitigate hallucinations from perspective of inference process, which does not require retraining the model. For example, \cite{zhang2024sled} modifies the logits in final layer by aggregating gradients from prior layers to improve consistency among decoder layers. \cite{chuang2023dola} utilizes contrastive decoding, which compares the differences in logits obtained from projecting the deep layers versus shallow layers to the vocabulary space, to reinforce outputs aligned with the factual patterns in particular decoder layers.

Previous research has indicated that deep layers of LLMs are prone to determine factual information, whereas shallow layers decode low-level representations such as \textit{was}, \textit{the}, and \textit{to} \citep{chuang2023dola}. Since hallucinations in LLMs are most frequently introduced in factual decoding, we identify that deep layers are the primary cause of this issue. To verify this, we conducted a preliminary experiment in which we progressively removed decoder layers from the LLaMA3-8B model, starting from the middle layer and moving toward the final layer. After each removal, we evaluated the model performance on the TruthfulQA \cite{lin2021truthfulqa} dataset. As shown in Figure \ref{fig:intro_acc}, skipping certain deeper layers can improve model performance, confirming that hallucinations are more likely to arise in deep decoder layers during inference.

To alleviate the hallucination problem and enhance the  reliability of LLMs, we introduced a novel decoding approach \textbf{DeLask} (\textbf{De}coder \textbf{La}yer \textbf{Sk}ipping) to dynamically partially skip decoder layers that are prone to generating hallucination, as illustrated in Figure \ref{fig:struc}. Our design is inspired by the theoretical finding of \cite{ahn2023transformers}, who found that a Transformer architecture is capable of implementing preconditioned gradient descent and that the \textit{\textbf{forward computation of an L-layer Transformer is conditionally equivalent to performing L steps of gradient descent}}. To define the \textbf{gradient}, we first adopt the early-exit strategy \citep{zhou2025efficiency}, projecting the hidden states of two consecutive layers directly onto the final linear classification head to obtain their probability distributions of logits. We then define the gradient from one decoding step to the next step as the difference between their probability distributions. To measure the directional change of the gradient across decoding steps, we further compute the cosine similarity between the gradients of two adjacent steps, which we refer to as the \textbf{driftance value}. A more negative driftance value, especially approching -1, indicates that the gradient direction at the later step deviates significantly from that of the earlier step, which can also be interpreted as the gradient shifting from a “descent” to an “ascent” direction. We regard such deviations as evidence that the current decoder layer is prone to generating hallucinations. Once a problematic layer is identified, we introduce a partial skipping mechanism that aggregates its hidden states with the previous one to attenuate its adverse influence. In contrast to fully skipping the layer—which can disrupt cross-layer coherence and destabilize the reasoning process—our approach maintains decoding consistency while effectively mitigating hallucinations. 

Extensive experiments across three task categories—factual hallucination detection, multiple-choice reasoning, and open-ended generation—on seven widely used benchmarks show that DeLask consistently enhances accuracy and robustness. These results highlight its effectiveness as a general framework for mitigating hallucinations and improving the reliability of large language models.

\section{Method}
\subsection{Decoding Process in LLMs}

Recent Large Language Models (LLMs) generally comprise an embedding layer, a series of $N$ transformer decoder blocks, and a final linear projection function $\phi(\cdot)$, which maps the hidden representations to a probability distribution over the vocabulary space. The token generation process for an $N$-layer LLM can be formally described as follows:

\begin{equation}\label{eq:layer-update}
\begin{cases}
h_{t}^{j+1} = h_{t}^{j} + \text{DecoderBlock}_{j}(h_{t}^{j}), & j = 0,\dots,N-1,\\[6pt]
p\bigl(x_{t+1}\mid x_{:t}\bigr)
  = \text{softmax}\bigl(\phi(h_{t}^{N})\bigr).
\end{cases}
\end{equation}

In Equation \eqref{eq:layer-update}, the term $\text{DecoderBlock}_{j}(\cdot)$ represents the transformations performed by the $j$-th decoder layer, which includes self-attention, feed-forward networks, and layer normalization. $h_{t}^{j}$ denotes the $t$-th token vector at the $j$-th layer. Each decoder layer incrementally update the previous hidden representations through a sequence of transformation, subsequently combining these refined representations with the original input via a residual connection. After progressing through all $N$ decoder layers, the resulting hidden state $h_{t}^{N}$ is fed into the linear projection $\phi(\cdot)$, producing logits for vocabulary tokens. Applying the softmax function to these logits yields a probability distribution for predicting the next token $x_{t+1}$. Although the linear projection $\phi(\cdot)$ is optimized exclusively based on the final-layer representation during training, its predictive capability heavily relies upon the incremental refinements produced by the intermediate layers.

\begin{figure}[t]
    \centering
    \begin{subfigure}[b]{0.49\linewidth}
        \centering
        \includegraphics[width=\linewidth]{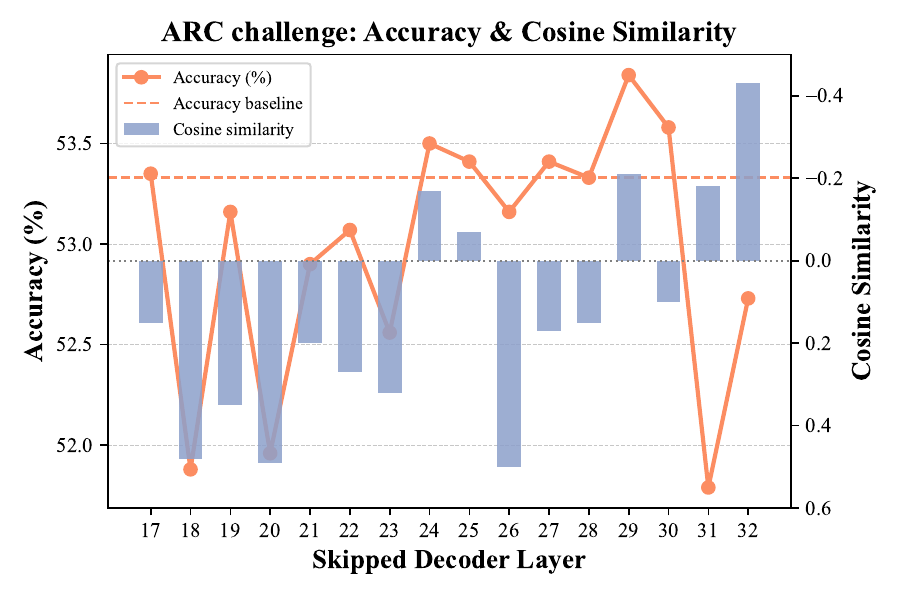}
        \caption{}
        \label{fig:motivation1}
    \end{subfigure}
    \begin{subfigure}[b]{0.49\linewidth}
        \centering
        \includegraphics[width=\linewidth]{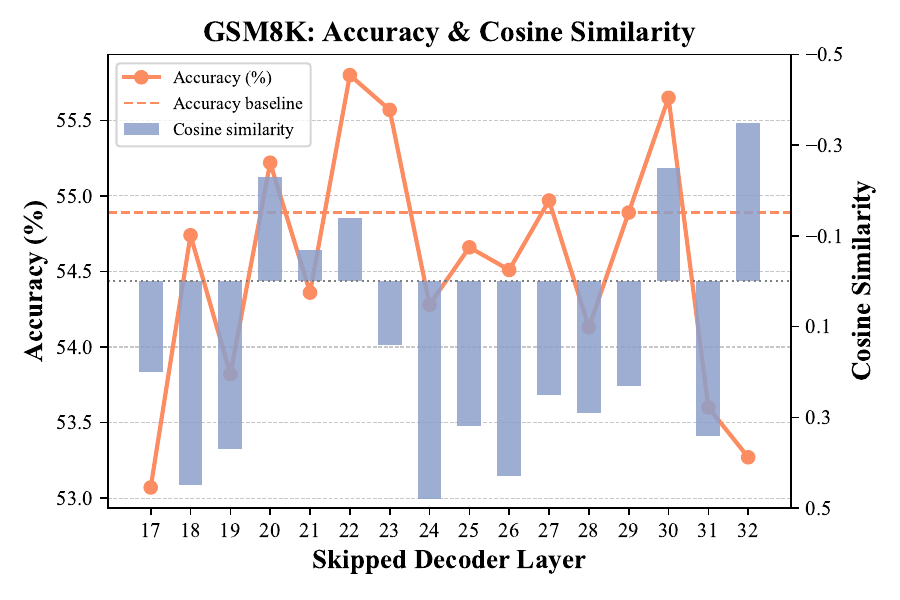}
        \caption{}
        \label{fig:motivation2}
    \end{subfigure}
    
    \caption{Relationship between accuracy and driftance values on ARC challenge \ref{fig:motivation1} and GSM8K \ref{fig:motivation2}: The bar plots show the cosine similarity between consecutive distribution differences (driftance values), while the line plots track model accuracy after skipping each decoder layer. Negative driftance values reflect substantial semantic drift between layers. Notably, layers with larger negative cosine similarity align with accuracy gains when skipped, suggesting that such layers are more prone to inducing hallucinations.}
    \label{fig:motivation}
    
\end{figure}

\subsection{Motivation}

We first conduct preliminary analysis using a 32-layer LLaMA3-8B model on ARC challenge \cite{cobbe2021training} and GSM8K \cite{clark2018think} datasets to validate the motivation behind our approach. Beginning from the middle layers, we sequentially skip decoder layers and evaluate the model’s accuracy after each removal. As shown in Figure \ref{fig:intro_acc}, omitting certain deeper layers improves overall performance, suggesting that these layers are more likely to introduce hallucinations and thereby harm output quality.

We then examine the correlation between this phenomenon and the driftance value. Specifically, we compute the driftance value for each layer and compare it with the model accuracy obtained after removing the corresponding decoder layer, as shown in Figure \ref{fig:motivation}. The results reveal that layers associated with negative driftance values tend to induce hallucinations, with the exception of the final layer. This exception arises because the final linear head is trained to rely heavily on the representation of the last decoder layer, and skipping it disrupts this alignment, leading to miscalibration. These findings demonstrate that driftance provides a reliable signal for detecting hallucination-prone layers and motivates its use in dynamic layer-skipping strategies to enhance LLM reliability.

\subsection{Hallucination Layer Detection}\label{sec:detection}

We first define the probability vector $\mathbf{p}_{t}^{(j)}$ which is sampled from hidden state at each intermediate layer $j$. To be specific, $\mathbf{z}_{t}^{(j)}$ is firstly given by implementing the linear activation $\phi(\cdot)$ to intermediate-layer hidden states was originally optimized using only the final-layer representation $h_{t}^{N}$, then we calculated the probability vector by the softmax operation:

\begin{equation}
\mathbf{z}_{t}^{(j)} = \phi(h_{t}^{j}), \quad
\mathbf{p}_{t}^{(j)} = \operatorname{softmax}\bigl(\mathbf{z}_{t}^{(j)}\bigr).
\end{equation}

Next, we calculate the difference in predictive distributions between adjacent layers:
\begin{equation}
\boldsymbol{\Delta}_{t}^{(j)} \;=\; \mathbf{p}_{t}^{(j)} - \mathbf{p}_{t}^{(j-1)}, \quad j=1,\dots,N.
\end{equation}
This vector precisely measures how the prediction of the next token evolves from one layer to the next, directly reflecting the implicit update direction in the model’s internal probability space.

To quantify directional consistency between successive layer updates, we define the \emph{driftance value} as cosine similarity between successive distribution differences:
\begin{equation}
c_{t}^{(j)} \;=\; 
\frac{\boldsymbol{\Delta}_{t}^{(j)}\cdot \boldsymbol{\Delta}_{t}^{(j-1)}}{\|\boldsymbol{\Delta}_{t}^{(j)}\|_2\,\|\boldsymbol{\Delta}_{t}^{(j-1)}\|_2},\quad j\geq 2.
\end{equation}

Here a negative driftance value indicates that the current layer actively diverges from earlier predictions, potentially causing the model to produce hallucinated outputs. To explicitly identify layers responsible for significant directional reversals and potential hallucinations, we introduce a negative threshold parameter \( \tau < 0 \). Specifically, we classify layer \( j \) as a hallucinated layer if the cosine similarity satisfies:
\begin{equation}
c_{t}^{(j)} \;<\; \tau.
\end{equation}
This threshold clearly distinguishes minor fluctuations from substantial directional reversals that are indicative of hallucination.

\subsection{Partial Layer Skipping}\label{sec:partial-skip}

To address hallucinations caused by layers identified in Section~\ref{sec:detection}, we propose a method termed \emph{Partial Layer Skipping}. Rather than completely skipping hallucinated layers or directly feeding earlier-layer hidden states into the final linear head \citep{zhou2025efficiency}, which may lead to feature misalignment.

Specifically, for each decoder layer \( j \) identified as hallucinated, we perform a controlled adjustment of the hidden state \( h_{t}^{j} \). We achieve this by blending it with the hidden state from the immediately preceding layer \( h_{t}^{j-1} \). Formally, we update the hidden state as follows:
\begin{equation}
h_{t}^{j} \leftarrow \alpha\,h_{t}^{j} + (1 - \alpha)\,h_{t}^{j-1}, \quad \alpha \in [0,1].
\end{equation}

Here, the coefficient \(\alpha\) controls the extent to which the representation from the previous layer is incorporated into the current hidden state. A larger value of \(\alpha\) corresponds to stronger reliance on the earlier layer representation, thus reducing more impact of the identified hallucination in current layer.

Preliminary experiments demonstrate that if \(c_{t}^{(j)}\) is closer to -1, the current layer is more likely to generate hallucination. To adaptively select an appropriate value of \(\alpha\), we employ a sigmoid mapping to ensure that \(\alpha\) remains within valid boundaries and responds appropriately to different levels of hallucination severity:

\begin{equation}
\alpha = \sigma\bigl(\beta(c_{t}^{(j)})\bigr), \quad \beta > 0.
\label{eq:7}
\end{equation}

In formula \ref{eq:7}, the sigmoid function inherently ensures numerical stability by bounding  within (0,1). Specifically, as the value of \(c_{t}^{(j)}\) decreases, \(\alpha\) approaches 0, effectively increasing the proportion of information from the preceding layer and thus bypassing the problematic current layer more significantly. Additionally, the scaling factor \(\beta\) allows us to easily adjust the sensitivity of \(\alpha\) to changes in \(c_{t}^{(j)}\), facilitating straightforward optimization. In particular, we summarize the overall procedure of DeLask in Algorithm~\ref{alg:partial-skip}. The skipping mechanism is applied selectively to the decoder layers in the range of the middle layer $N/2{-}1$ to $N{-}2$. 

\begin{algorithm}[htbp]
\caption{Overall Process of DeLask}\label{alg:partial-skip}
\begin{algorithmic}
\REQUIRE Hidden states $\{h_t^j\}_{j=1}^N$, linear head $\phi$, threshold $\tau<0$, sensitivity $\beta>0$
\ENSURE Adjusted hidden states for final decoding
\FOR{$j = \lfloor N/2\rfloor -1 ,\dots,N-2$}
  \STATE $\mathbf{p}_t^{(j-1)} \leftarrow \mathrm{softmax}\bigl(\mathbf\phi(h_t^{(j-1)})\bigr)$
  \STATE $\Delta_t^{(j)} \leftarrow \mathbf{p}_t^{(j)} - \mathbf{p}_t^{(j-1)}$, $\Delta_t^{(j-1)} \leftarrow \mathbf{p}_t^{(j-1)} - \mathbf{p}_t^{(j-2)}$
  \STATE $c_t^{(j)} \leftarrow \dfrac{\Delta_t^{(j)} \cdot \Delta_t^{(j-1)}}{\|\Delta_t^{(j)}\|_2\;\|\Delta_t^{(j-1)}\|_2}$
  \IF{$c_t^{(j)} < \tau$}
    \STATE $\alpha \leftarrow \sigma\bigl(\beta\,c_t^{(j)}\bigr)$
    \STATE $h_t^j \leftarrow \alpha\,h_t^j + \bigl(1 - \alpha\bigr)\,h_t^{j-1}$
  \ENDIF
\ENDFOR
\STATE \textbf{return} $\{h_t^j\}_{j=1}^N$ \COMMENT{for final projection and decoding}
\end{algorithmic}
\end{algorithm}

\vspace{-0.4cm}

\section{Experiments}

 We evaluated DeLask against five representative decoding strategies: regular, DOLA \cite{chuang2023dola}, ITI\cite{li2023inference}, UAD \cite{timor2025accelerating}, and Activation Decoding (AD) \cite{chen2024context} on both LLaMA2-7B and LLaMA3-8B. These methods have been widely explored for mitigating hallucinations in large language models, ranging from simple logits calibration to uncertainty- and activation-based decoding. Together, they form a diverse and competitive set of baselines, enabling a comprehensive assessment of DeLask across factual hallucination, multiple-choice, and open-ended generation tasks. All parameter configurations were based on the default settings provided by the open-source LLM evaluation framework, \texttt{lm-evaluation-harness}. The batch size was fixed at 1 for all tasks to ensure consistency across experiments.

\begin{table*}[t]
  \centering
  \renewcommand{\arraystretch}{1}
  \caption{Experimental results of two LLaMA variants across different decoding methods on 
  (i) factual hallucination benchmarks (TruthfulQA and FACTOR), 
  (ii) multiple-choice tasks (MMLU and ARC-C), 
  and (iii) open-ended generation tasks (TriviaQA, GSM8K, and CoQA).}
  \resizebox{\textwidth}{!}{%
  \begin{tabular}{ll|cccc|cc|ccc}
    \toprule
    \multirow{2}{*}{\bfseries Models} & 
    \multirow{2}{*}{\bfseries Decoding} &
      \multicolumn{4}{c|}{\bfseries Factual Hallucination} &
      \multicolumn{2}{c|}{\bfseries Multiple Choice} &
      \multicolumn{3}{c}{\bfseries Open-ended Generation} \\
    \cmidrule(lr){3-6} \cmidrule(lr){7-8} \cmidrule(lr){9-11}
    & & \bfseries TruthQA-MC1 & \bfseries TruthQA-MC2 & \bfseries FACTOR-News & \bfseries FACTOR-Wiki & 
        \bfseries MMLU & \bfseries ARC-C &
        \bfseries TriviaQA & \bfseries GSM8K & \bfseries CoQA \\
    \midrule
    \multirow{6}{*}{LLaMA3-8B}
      & Regular                 & 39.72 & 57.10 & 67.47 & 57.77 & 65.25 & 53.33 & 71.61 & 54.89 & 67.40 \\
      & DOLA                    & 38.43 & 56.38 & 65.12 & 54.47 & 61.36 & 50.34 & 70.77 & 54.74 & 62.91 \\
      & ITI & 37.14 &57.05 & 67.22 & 57.71 & 64.28 & 52.49 & 67.37 & 54.36 & 67.71 \\
      & UAD                      & 39.59 & 57.44 & 67.55 & 56.81 & 65.43 & 53.72 & 64.09 & 55.02 & 64.67 \\
      & AD     & 35.45 & 55.74 & 62.15 & 53.72 & 60.69 & 48.93 & 70.48 & 53.39 & 63.40 \\
      & \textbf{DeLask}          & \textbf{41.74} & \textbf{58.76} & \textbf{68.56} & \textbf{58.71} & \textbf{65.43} & \textbf{54.10} & \textbf{72.65} & \textbf{55.42} & \textbf{68.05} \\
    
    \midrule
    \multirow{6}{*}{LLaMA2-7B}
      & Regular                 & 33.51 & 51.12 & 64.82 & 55.77 & 45.54 & 45.98 & 63.34 & 14.70 & 62.55 \\
      & DOLA                    & 33.71 & \textbf{53.17} & 65.01 & 56.51 & 43.57 & 45.63 & 62.50 & 13.34 & 60.76 \\
      & ITI & 33.39 & 52.03 & 65.17 & 56.21 & 45.85 & 44.13 & 63.89 & 14.74 & 62.48 \\
      & UAD                      & 32.01 & 52.47 & 64.46 & 55.89 & 45.24 & 43.17 & 54.09 & 14.86 & 61.65 \\
      & AD    & 34.14 & 52.03 & 65.59 & 56.71 & \textbf{45.90} & 46.07 & 63.57 & 13.76 & 62.38 \\
      & \textbf{DeLask}          & \textbf{35.02} & 52.27 & \textbf{65.95} & \textbf{57.85} & 45.47 & \textbf{46.24} & \textbf{64.32} & \textbf{15.04} & \textbf{63.76} \\
    \bottomrule
  \end{tabular}%
  \label{tab:decoding-comparison}
  }
\end{table*}



\vspace{-0.2cm}

\subsection{Experiment Results on Factual Hallucination Task}
We evaluate DeLask on TruthfulQA \cite{lin2021truthfulqa}, which tests factual consistency under adversarial prompts, and FACTOR \cite{muhlgay2023generating}, which measures reliability across news and Wikipedia domains. As shown in Table \ref{tab:decoding-comparison}, DeLask achieves the strongest gains on LLaMA3-8B, improving TruthfulQA-MC1/MC2 accuracy to 41.74\%/58.76\% and FACTOR-News/Wiki accuracy to 68.56\%/58.71\%, consistently outperforming Regular and nearly all baselines.

\subsection{Experiment Results on Multiple Choice Task}
We evaluate on MMLU \cite{hendrycks2020measuring}, which covers diverse academic and professional domains, and ARC-Challange (ARC-C) \cite{clark2018think}, which requires complex scientific reasoning. On LLaMA2-7B, several baselines show moderate gains, but their effectiveness diminishes on LLaMA3-8B. In contrast, DeLask maintains improvements, reaching 65.43\% on MMLU and 54.10\% on ARC-C, highlighting its robustness on stronger models.

\subsection{Experiment Results on Open-ended Generation Task}
We evaluate on TriviaQA \cite{joshi2017triviaqa}, a reading comprehension dataset; GSM8K \cite{cobbe2021training}, which involves multi-step mathematical reasoning; and CoQA \cite{reddy2019coqa}, a conversational QA benchmark. DeLask consistently surpasses baselines on LLaMA3-8B, 55.42\% on GSM8K, and 68.05\% on CoQA, while yielding steady improvements on LLaMA2-7B. These results confirm its effectiveness on diverse open-ended tasks.

\section{Ablation Studies}
\subsection{Evaluation of Hyperparameter Sensitivity}
To evaluate the hyperparameter sensitivity of DeLask, we conduct an ablation study on the skipping threshold $\tau$ and the scaling factor $\beta$ using LLaMA3-8B, with results reported on TruthfulQA in Table~\ref{tab:tau-beta-ablation}. For the skipping threshold $\tau$, DeLask consistently improves over the baseline, with the best results at $\tau=-0.15$. Similarly, for the scaling factor $\beta$, most configurations yield higher accuracy than the baseline, with $\beta=0.01$ reaching the same peak and $\beta=0.005$ also achieving 40.37\% on MC1 and 58.25\% on MC2. These results indicate that although the thresholds and scaling factors are selected manually, DeLask consistently outperforms the baseline under most configurations, demonstrating both stability and effectiveness.
\begin{table}[htbp]
  \centering
  \small
  \renewcommand{\arraystretch}{0.8}
  \caption{Experimental results on TruthfulQA under different thresholds $\tau$ (top) and scaling factors $\beta$ (bottom) using LLaMA3-8B.}
  \label{tab:tau-beta-ablation}
  \resizebox{0.5\textwidth}{!}{%
  \begin{tabular}{cc|ccccc}
    \toprule
    \multicolumn{7}{c}{\textbf{Skipping threshold $\tau$}} \\
    \midrule
    \textbf{Metric} & \textbf{Baseline} & $-0.25$ & $-0.20$ & $-0.15$ & $-0.10$ & $-0.05$ \\
    \midrule
    MC1 & 39.72 & 39.91 & 40.58 & 41.74 & 40.28 & 40.92  \\
    MC2 & 57.10 & 57.59 & 57.79 & 58.76 & 57.15 & 56.91 \\
    \midrule
    \multicolumn{7}{c}{\textbf{Scaling factor $\beta$}} \\
    \midrule
    \textbf{Metric} & \textbf{Baseline} & $0.1$ & $0.05$ & $0.01$ & $0.005$ & $0.001$ \\
    \midrule
    MC1 &39.72 & 39.92 & 41.14 & 41.74 & 40.37 & 41.12 \\
    MC2 & 57.10 & 56.82 & 57.63 & 58.76 & 58.25 & 56.62 \\
    \bottomrule
  \end{tabular}%
  }
\end{table}

\subsection{Evaluation of Different Start Layer}
We evaluate the effect of different starting layers of DeLask on LLaMA3-8B across five datasets spanning open-ended generation and reasoning tasks, as shown in Table \ref{tab:layer-start-ablation}. Applying it from shallow layers leads to lower performance, as early representations are less semantically stable and more sensitive to pruning. In contrast, starting from mid layers achieves the best results, with layer 16 yielding peak performance across all tasks, while later activation brings only marginal gains.

\begin{table}[htbp]
  \centering
  \small
  \setlength{\tabcolsep}{4pt}      
  \renewcommand{\arraystretch}{0.8} 
  \caption{Experimental results of different starting layer of DeLask on 1) multiple-choice datasets: MMLU and ARC-C, and 2) open-ended generation tasks: TriviaQA, GSM8K, and CoQA using LLaMA3-8B.}
  \label{tab:layer-start-ablation}
  \begin{tabular}{c|ccccc}
    \toprule
    \bfseries Start Layer & \bfseries MMLU & \bfseries TriviaQA & \bfseries ARC-C & \bfseries GSM8K & \bfseries CoQA \\
    \midrule
      4  & 60.17 & 63.76 & 46.84 & 37.24 & 62.89 \\
      8  & 62.48 & 68.43 & 49.95 & 45.31 & 64.23 \\
     12  & 65.28 & 70.58 & 52.13 & 53.14 & 67.63 \\
     16  & 65.43 & 72.65 & 54.10 & 55.42 & 68.05 \\
     20  & 65.14 & 71.52 & 53.74 & 54.72 & 67.92 \\
     24  & 65.39 & 71.37 & 53.43 & 55.02 & 67.76 \\
    \bottomrule
  \end{tabular}
\end{table}

\section{Conclusion}
In this paper, we addressed the challenge of hallucinations in Large Language Models (LLMs), which critically limit their reliability in high-stakes applications. Building on the observation that deep decoder layers are a primary source of hallucinations, we introduced \textbf{DeLask}, a decoding framework that dynamically skips problematic layers based on gradient driftance. By partially aggregating hidden states from consecutive layers, DeLask preserves cross-layer consistency while reducing erroneous outputs. Grounded in the theoretical link between Transformer computation and gradient descent, our method offers a principled and efficient alternative to retraining-based approaches. Extensive experiments across multiple LLMs and benchmarks confirm that DeLask consistently mitigates hallucinations and enhances overall reliability, highlighting its potential as a lightweight and generalizable solution for improving the robustness of large-scale language models.

\clearpage

\bibliographystyle{IEEEbib}
\bibliography{icassp}

\end{document}